\documentclass[9pt]{worldcomp}
\title{mpEAd: Multi-Population EA Diagrams}

\usepackage{graphicx}
\usepackage{float}
\usepackage{wrapfig}
\usepackage{array}
\usepackage{subfig}
\usepackage{fixltx2e}
\usepackage[sorting=none,style=numeric,backend=bibtex]{biblatex}
\author{
	{\bfseries Sebastian Lenartowicz$^1$ and Mark Wineberg$^2$}\\
	University of Guelph\\50 Stone Road, Guelph ON N1G 2W1\\
	$^1$\textit{slenarto@mail.uoguelph.ca}\\$^2$\textit{mwineber@uoguelph.ca}
}

\begin{document}
	\maketitle

	\begin{abstract}
		Multi-population evolutionary algorithms are, by nature, highly complex and difficult to describe. Even two populations working in concert (or opposition) present a myriad of potential configurations that are often difficult to relate using text alone. Little effort has been made, however, to depict these kinds of systems, relying solely on the simple structural connections (related using \textit{ad hoc} diagrams) between populations and often leaving out crucial details. In this paper, we propose a notation and accompanying formalism for consistently and powerfully depicting these structures and the relationships within them in an intuitive and consistent way. Using our notation, we examine simple co-evolutionary systems and discover new configurations by the simple process of ``drawing on a whiteboard''. Finally, we demonstrate that even complex, highly-interconnected systems with large numbers of populations can be understood with ease using the advanced features of our formalism.
	\end{abstract}

	\section{Introduction}
	\label{sec:intro}
		From the beginning, it has been obvious that evolutionary algorithms (EAs) could make use of multiple populations in order to facilitate more complex searches and increase the power of the search itself. Though it is doubtless that others exist, there are four main types of multi-population systems that have been investigated. In the island model\cite{whitley1999island}, solutions move between different discrete populations that use the same objective function. The predator/prey model\cite{hillis1990co} uses multiple populations to perform fitness evaluation -- an individual in one population is compared against one or more individuals in another population, where its fitness score increases as those of the others decrease (and vice versa). Co-operative co-evolution\cite{potter1994cooperative} is yet another different system, in which members of each different population comprise different elements of the complete solution (and members must be drawn from each population to form and evaluate a full solution). Finally, hierarchical systems, though sparsely investigated (and with conflicting definitions\cite{sefrioui2000hierarchical}\cite{gulsen1999hierarchical}), use a variety of different multi-population structures that utilize levels in order to achieve their aims.

		Investigation into multi-population EA systems has waxed and waned over the decades, likely because such systems tend to produce complexities that simple single-population systems do not incur. Populations may exchange both genetic and evaluative information\cite{whitley1999island}\cite{de2006evolutionary}, and, in more esoteric systems, other types of information as well\cite{li2002multilevel}. There are often complex spatial structures connecting the populations formed by this web of relationships\cite{whitley1999island}, and the structure may be distinctly different depending upon the nature of the information\cite{li2002multilevel}. Furthermore, there are recursive effects between multiple populations -- as seen in many co-evolutionary systems\cite{potter1994cooperative}\cite{de2006evolutionary}\cite{wiegand2004spatial} -- that lead to problems such as Red Queen\cite{pagie2000information}, the loss-of-gradient effect\cite{wiegand2004spatial}, and decoupling\cite{wiegand2004spatial}. This complexity in the structure of information flow throughout the system is in addition to the actual movement of individuals between populations, as seen in island-model migration\cite{whitley1999island}. All of this is further exacerbated when considering hierarchical multi-population systems. Confusion about the very idea of what constitutes a hierarchical system can easily be seen in even a cursory review of the literature\cite{gulsen1999hierarchical}\cite{li2002multilevel}\cite{man2012genetic}; all of these systems are called hierarchical and incorporate elements of hierarchy, but all are different, with very few elements in common!

		To combat the confusion arising from all this complexity, we have developed a graphical formalism that encapsulates the different relationships that can exist between multiple populations in an EA system. The multi-population EA diagram (mpEAd, pronounced ``emm-pede'') employs a concise visual grammar to depict multiple populations and the information flow between them, in a similar way to how the Unified Modelling Language (UML)\cite{rumbaugh2004unified} captures the categorization of and relationships between the component parts of object-oriented software systems.

		At this juncture, it is important to note that the mpEAd system is not intended to depict the internal mechanisms and dynamics of single populations; in other words, it does not describe the different selection methods, reproductive operators, etc., of the evolutionary algorithm used by a population. Instead, mpEAd treats each population as a black box, accepting inputs and producing outputs for consumption by other populations or itself. This is approximately analogous to the static ``structural diagrams'' of UML, which depict the relationship between the system components (such as classes and subsystems) rather than their time-dependent activities, which are modelled in its ``behavioural'' diagrams. It should also be noted, however, that, while inspired by UML, mpEAd does not incorporate any of the notational conventions found in it, instead using a visual language that is more suited to modelling EA systems. Finally, it is important to stress that mpEAd is more than the topological structure of the populations, as frequently seen when discussing migration. The relationships depicted in mpEAd are much broader in scope, and model all types of information exchange between populations, with migration being only a single subset.

		This paper is divided into two parts: the first presenting the basic elements of mpEAd necessary to structure any system and the second extending this notation to model more complex systems. Examples are provided throughout to demonstrate the notation and power of the mpEAd system.

	\section{Essential Structure of mpEAd}
	\label{sec:struct}
		All attempts to depict multi-population EAs base themselves on the graph formalism, with populations represented as nodes (with edges taking variable meanings). However, as with class diagrams in UML, where nodes are classes and edges the relationships between them (messages, roles, etc.), the complexity of an EA system is not fully captured by a simple graph and additional diagrammatic formats are required in order to capture all aspects of the EA's functioning.

		With this in mind, a few design principles emerged organically while constructing the mpEAd formalism: intuitiveness, consistency, distinctiveness, and simplicity. While the roles of intuitiveness and simplicity are simple and intuitive (and yet sometimes difficult to achieve), a brief explanation of the other principles is warranted. While consistency may seem similarly straightforward, the notion of establishing a visual grammar is often lost when attempting to communicate information. Ensuring that similar things appear similar helps bring to mind meaning and allows ease of learning, extensibility, and creativity. In contrast to this, differences should appear different. This promotes readability and ease of interpretation. We also held to two supplementary principles: that the diagram should render naturally in black and white (for publication), and that it should be simple to draw by hand on a whiteboard or piece of paper.

		\subsection{Basic Elements of mpEAd}
		\label{ssec:basics}
			The graph at the heart of mpEAd incorporates two types of nodes: population nodes and computation nodes, as well as a number of different edge types. All of these are described in detail below.
			
			\subsubsection{Population Nodes}
				A population node corresponds to a single optimization algorithm (usually an EA) and a set of solutions (the population). It is denoted using a simple hollow rectangle with solid borders and the name of the node written inside. The population node, when drawn, always includes a set of multiple parallel lines, which serve as a visual reminder of the many members of the population inside.

			\subsubsection{Information as Edges}
\begin{figure}
					\centering
					\begin{tabular}{|m{0.1\textwidth}|m{0.3\textwidth}|} \hline
						Depiction & Type of Information \\ \hline
						\vspace{5pt}\includegraphics[width=0.1\textwidth,keepaspectratio=true]{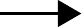} & Phenotypic \\ \hline
						\vspace{5pt}\includegraphics[width=0.1\textwidth,keepaspectratio=true]{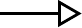} & Genotypic \\ \hline
						\vspace{5pt}\includegraphics[width=0.1\textwidth,keepaspectratio=true]{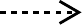} & Evaluative \\ \hline
					\end{tabular}
					\caption{Arrowheads and edges used in mpEAd.}
					\label{fig:arrows}
				\end{figure}

				Edges in the mpEAd graph are used to model the information flow between nodes. As information flow is directional, mpEAd becomes a directed graph, and, per the convention, uses arrows to indicate direction. The types of information available in a multi-population EAs are numerous and varied but can, however, be categorized into two distinct groups: genetic and non-genetic. Genetic information consists of any information used to construct or embody a solution, being referred to as genotypic and phenotypic, respectively. These are often interchangeable, as the genotypic is often immediately evaluated for fitness as if it were a phenotype. Furthermore, even when phenotypes are used, they are produced and consumed during a single decoding/evaluation step, and discarded upon completion. This often stops being true in multi-population systems, where phenotypes may be passed around for various purposes and used by multiple evaluations. The flow of genetic information is represented using a solid edge with a closed arrowhead that is either hollow (genotypic information) or filled (phenotypic information). Examples of these edges are given in Figure \ref{fig:arrows}.

				Evaluative information, such as objective and fitness values, is nearly universal in optimization; this information is represented in mpEAd using a dashed edge with an open arrowhead, with the dashed edge serving to indicate that evaluative information is non-genetic in nature. While other types of non-genetic information, such as statistical or control information, may exist in multi-population optimizing systems, a full discussion of these is outside the scope of this paper.

				 If the same unit of information has multiple simultaneous recipients, the edge is drawn as diverging from a single source. Conversely, if two pieces of information are required, the lines to the node are kept separate and distinct and are not combined in a similar way.

			\subsubsection{Computation Nodes}
				The second type of node, the computation node, is less obvious and is one of the elements that makes the mpEAd formalism more than simply a topological model of the connections between populations. The role of the computation node is to take in one or more streams of information, perform processing on them, and to provide the result to another node or nodes. Computation nodes perform a variety of information processing operations, including but not limited to decoding genotypic information into phenotypic information, evaluating fitness, and combining information from different sources. The computation node is depicted using a large hollow circle, often labelled with a name, such as the name of the fitness function used for evaluation.
				
				In general, the border of the circle matches the line type of the output edge(s). If the output is of mixed type, the circle's border alternates evenly between solid and dashed. It should be noted that the fundamental difference between a computation node and a population node is that, while both types can perform information processing, the computation node is stateless and does not store information, only taking input and producing output based upon it.

			\begin{figure}
				\centering
				\subfloat[
					A basic single-population EA solving the 1max problem.
				]{
					\hspace{0.05\textwidth}
					\includegraphics[width=0.1\textwidth,keepaspectratio=true]{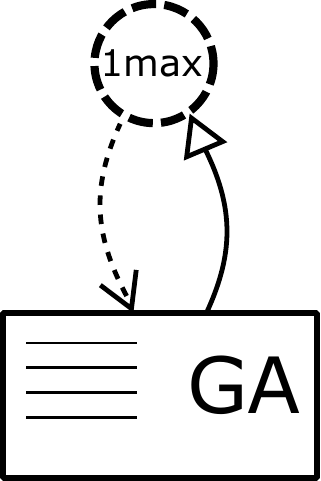}
					\hspace{0.05\textwidth}
					\label{sfig:basic-shorthand}
				}
				\hspace{5pt}
				\subfloat[
					A variant that decodes genotypes into phenotypes for evaluation.
				]{
					\hspace{0.05\textwidth}
					\includegraphics[width=0.1\textwidth,keepaspectratio=true]{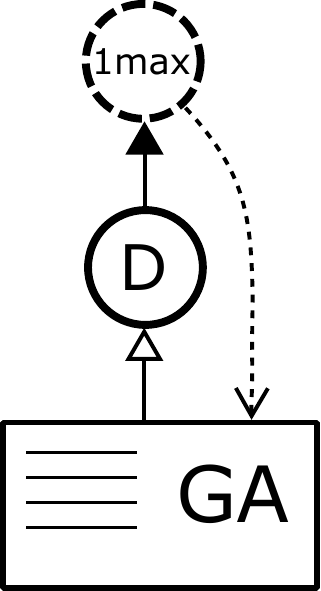}
					\hspace{0.05\textwidth}
					\label{sfig:basic-stats}
				}
				\caption{Examples of a basic EA.}
				\label{fig:basics}
			\end{figure}

			\subsubsection{Putting the Basic Elements Together}
				The simplest multi-population system is one with only a single population; i.e. the simple EA. Two simple examples of this kind of system, solving the universally-known 1max problem, can be seen in Figure \ref{fig:basics} to provide context for understanding the basic elements of mpEAd. In Figure \ref{sfig:basic-shorthand}, the genetic information is evaluated directly, with the resulting fitness value being passed back to be stored within the population. In Figure \ref{sfig:basic-stats}, the same evaluation takes place, but must first be decoded into a phenotype before evaluation can take place.

		\subsection{mpEAd in Action}
			\begin{figure}
				\centering
				\includegraphics[height=0.06\textwidth,keepaspectratio=true]{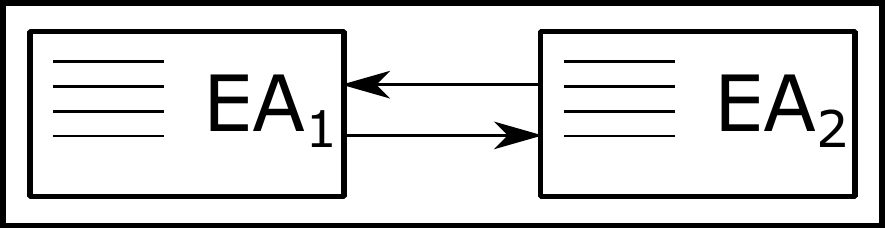}
				\caption{A typical na\"ive way to model the systems in Figure \ref{fig:coev}.}
				\label{fig:stupid-basic}
			\end{figure}

\begin{figure}
				\centering
				\subfloat[
					A basic predator/prey co-evolutionary system.
				]{
					\includegraphics[width=0.19\textwidth,keepaspectratio=true]{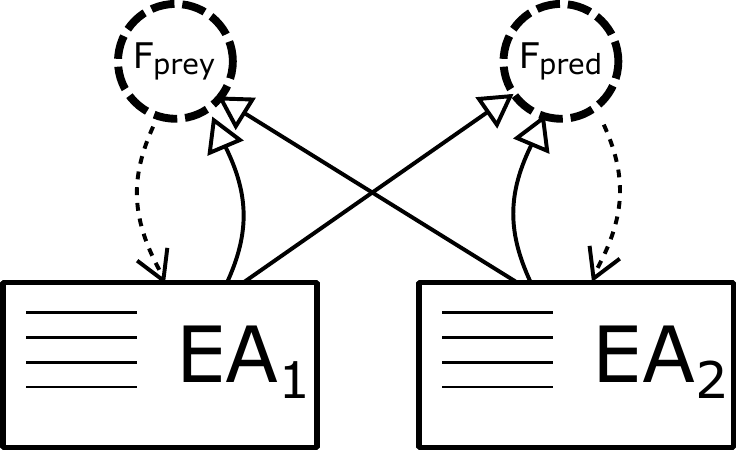}
					\label{sfig:pred-prey}
				}
				\hspace{0.025\textwidth}
				\subfloat[
					Standard co-operative co-evolution.
				]{
					\includegraphics[width=0.19\textwidth,keepaspectratio=true]{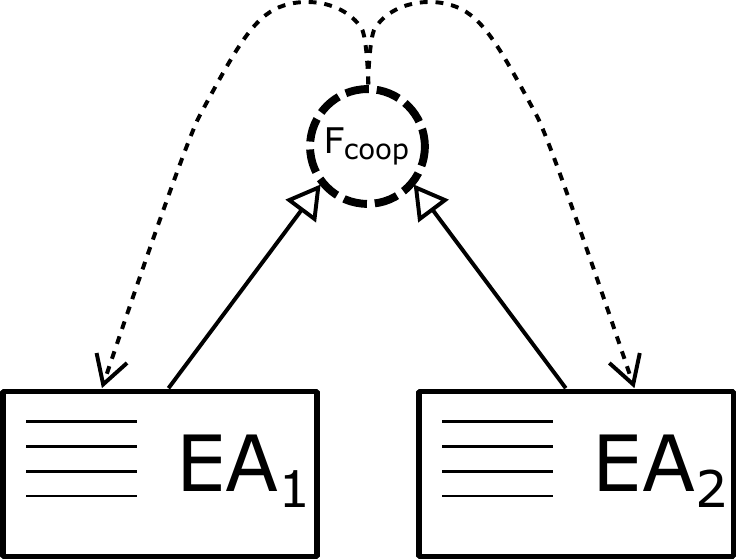}
					\label{sfig:coop}
				}

				\vspace{6pt}
				\subfloat[
					A hybrid of co-operative and predator/prey co-evolution with modification.
				]{
					\includegraphics[width=0.19\textwidth,keepaspectratio=true]{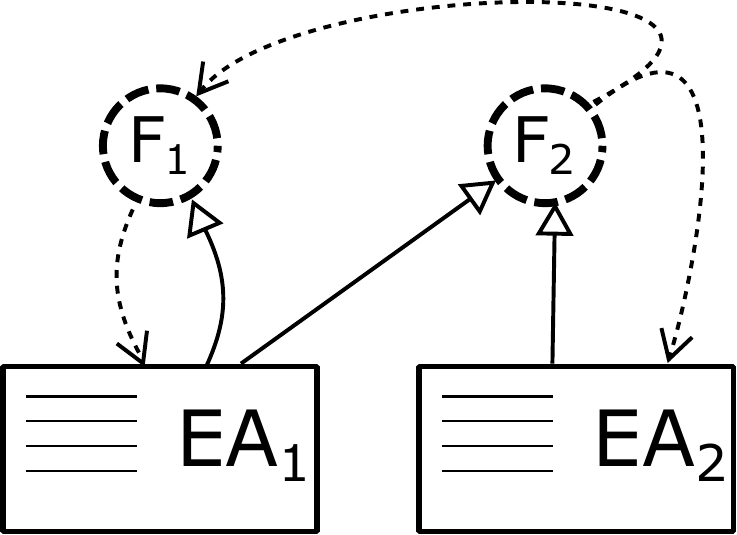}
					\label{sfig:coop-mod}
				}
				\hspace{0.025\textwidth}
				\subfloat[
					A hybrid of co-operative and predator/prey co-evolution with cross-modification.
				]{
					\includegraphics[width=0.19\textwidth,keepaspectratio=true]{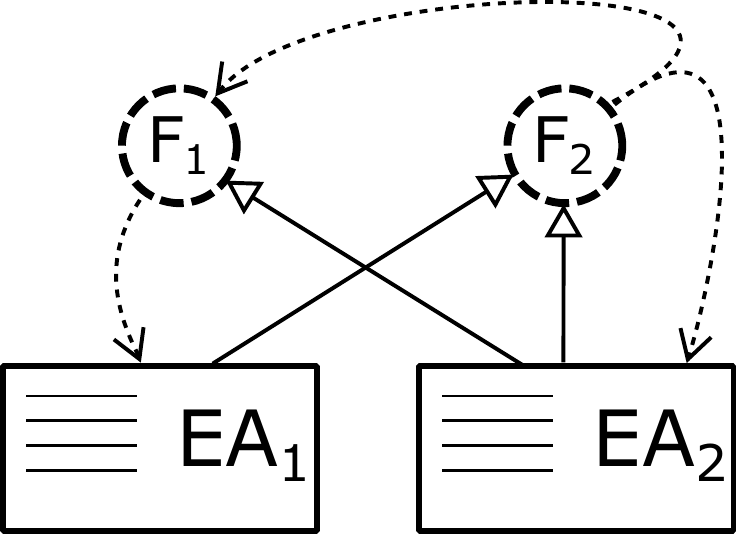}
					\label{sfig:coop-cross-mod}
				}
				\caption{Examples of various co-evolutionary systems. The two complementary systems in \ref{sfig:coop-mod} and \ref{sfig:coop-cross-mod} are both, to the best of our knowledge, novel.}
				\label{fig:coev}
			\end{figure}

				While efforts have been made in the past to model the interactions between populations, they are often simplistic and rely on \textit{ad hoc} notations, similar to what is seen in Figure \ref{fig:stupid-basic}. The power of mpEAd becomes apparent in comparison to this, as it permits much more accurate and detailed modelling of how the populations interact. All of the diagrams in Figure \ref{fig:coev} are different co-evolutionary systems that would be equivalent to the one in Figure \ref{fig:stupid-basic}. Many disparate types of multi-population systems (in this case, a variety of co-operative and competitive co-evolutionary systems) can thus be represented in a way such that their similarities, as well as their differences, become apparent.

			Figures \ref{sfig:pred-prey} and \ref{sfig:coop} depict a pair of standard co-evolutionary systems that are familiar to most EC researchers. In Figure \ref{sfig:pred-prey}, the diagram shows both EA populations sending members to the predator and prey evaluation functions, which are used to compute the two different fitnesses. Figure \ref{sfig:coop}, in comparison, depicts a co-operative system -- where the individuals from the two populations are combined to produce a single fitness value that is applied to both. 

			With the co-evolutionary systems in Figures \ref{sfig:coop-mod} and \ref{sfig:coop-cross-mod}, interesting possibilities begin to appear. These systems are unknown in the literature, but by using mpEAd can easily be conceived of, modelled, and constructed. On examination, they appear to be a hybrid between the co-operative and predator/prey systems seen in Figures \ref{sfig:pred-prey} and \ref{sfig:coop}. For both of these systems, a single fitness value is produced based on input from the two different populations; however, for one of the populations, the fitness value is modified, either by the individual itself or the individual from the other population. Many different co-evolutionary systems can be easily constructed in an analogous manner, demonstrating the power of the mpEAd formalism for both modelling and discovery.

			\section{Advanced Features of mpEAd}
			\label{sec:advanced}
				While mpEAd has many additional, powerful features that make modelling even very complex systems trivial, there are too many to exhaustively discuss here. Instead, we concentrate on the ones necessary to provide the most understanding for the most systems. To this end, we describe four additional features: edge labels, inset arrowheads, macro boxes, and ellipsis notation.

				\subsection{Edge Labels}
					The co-evolutionary examples given in Figure \ref{fig:coev} hide a great deal of important detail regarding the structure of the information being passed around the system. In particular, when considering evaluation, they do not provide information about how many individuals are required, where they come from, how they are to be selected, where they are to be stored, etc. There are many approaches for matching individuals between co-evolutionary populations, ranging from simple pairing, to pairing with an elite, to exhaustive combination pairing. Because these different constructs would result in an mpEAd that otherwise looks the same, edge labels can be used to disambiguate the selection and matching of individuals between nodes.

					The simplest kind of edge label is a letter variable, which is used to indicate sequential iteration through the individuals in a given population, both for selection and storage of incoming values. These variables can be thought of as indices to individuals within the population. Similarly, numbers (either single or in a range) are used to indicate when multiple individuals are drawn from a population in order to perform a computation. The algorithm by which these individuals may be chosen can be specified by a forward slash followed by an algorithm name or symbol (which should be described in accompanying literature) following the number or range. A more thorough discussion of this notation is outside the scope of this paper and will be explored in future work. An asterisk (*) is a special case of numeric value, in which the entire population is used and, in this case, no algorithm specifier may be provided. The asterisk was chosen rather than the more conventional $n$ used in computer science because $n$ could be mistaken for an index variable, and the asterisk as a symbol is commonly used to denote ``everything'' or ``all'' (e.g. as used in string matching and the Unix command line).

			\begin{figure}
				\centering
				\subfloat[
					The basic predator/prey co-evolutionary system from Figure \ref{sfig:comparative-basic} with edge labels.
				]{
					\includegraphics[width=0.19\textwidth,keepaspectratio=true]{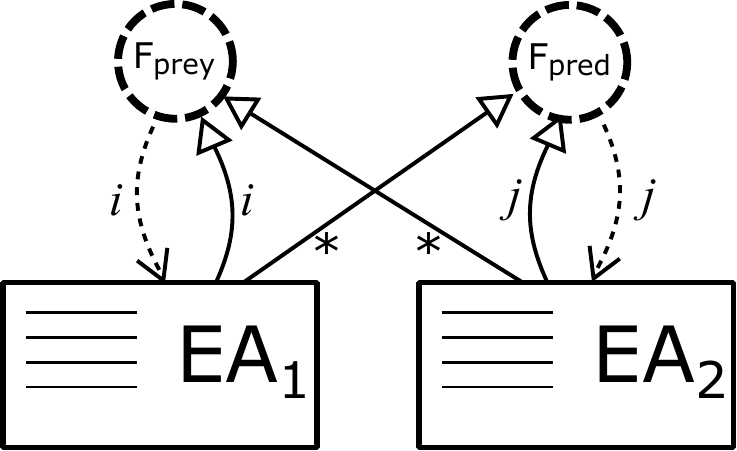}
					\label{sfig:pred-prey-multi}
				}
				\hspace{0.025\textwidth}
				\renewcommand{\thesubfigure}{d}
				\subfloat[
					The standard co-operative co-evolution from Figure \ref{sfig:coop} with edge labels.
				]{
					\includegraphics[width=0.19\textwidth,keepaspectratio=true]{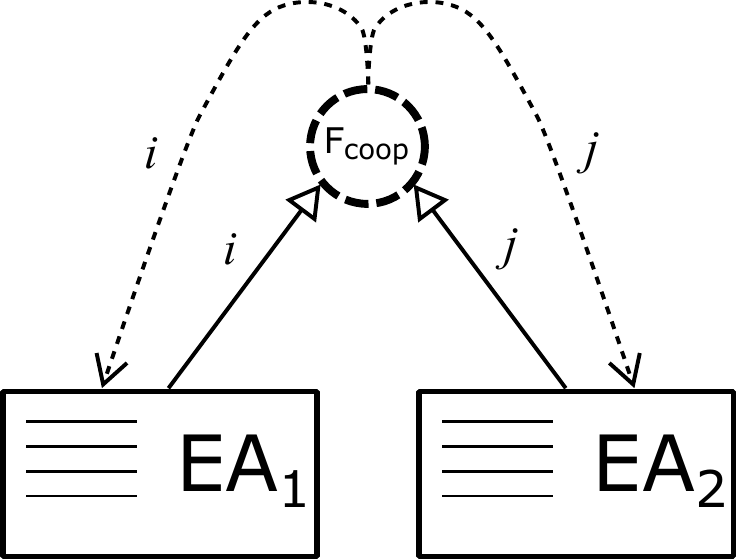}
					\label{sfig:coop-multi}
				}

				\vspace{6pt}
				\renewcommand{\thesubfigure}{b}
				\subfloat[
					A variant of the basic predator/prey system.
				]{
					\includegraphics[width=0.19\textwidth,keepaspectratio=true]{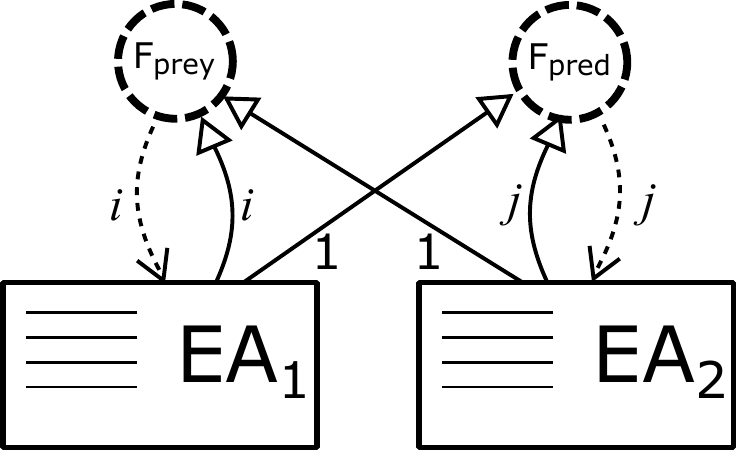}
					\label{sfig:pred-prey-multi-single}
				}
				\hspace{0.025\textwidth}
				\renewcommand{\thesubfigure}{e}
				\subfloat[
					A slightly different co-operative system that appears similar to a predator/prey system.
				]{
					\includegraphics[width=0.19\textwidth,keepaspectratio=true]{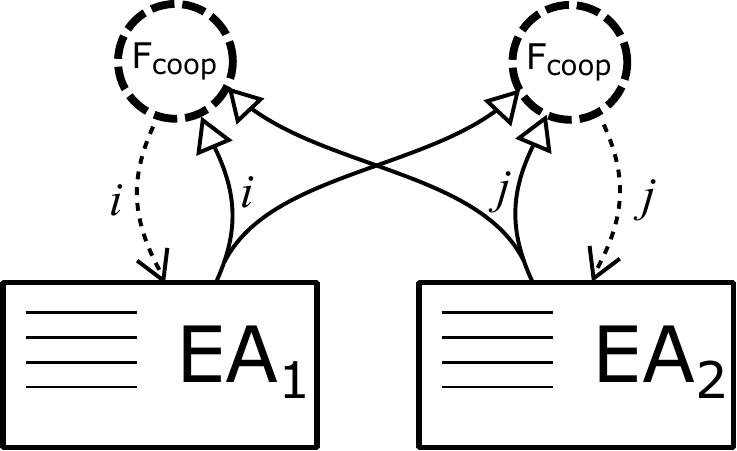}
					\label{sfig:comparative-coop}
				}

				\vspace{6pt}
				\renewcommand{\thesubfigure}{c}
				\subfloat[
					A more complex variant of the predator/prey system.
				]{
					\includegraphics[width=0.19\textwidth,keepaspectratio=true]{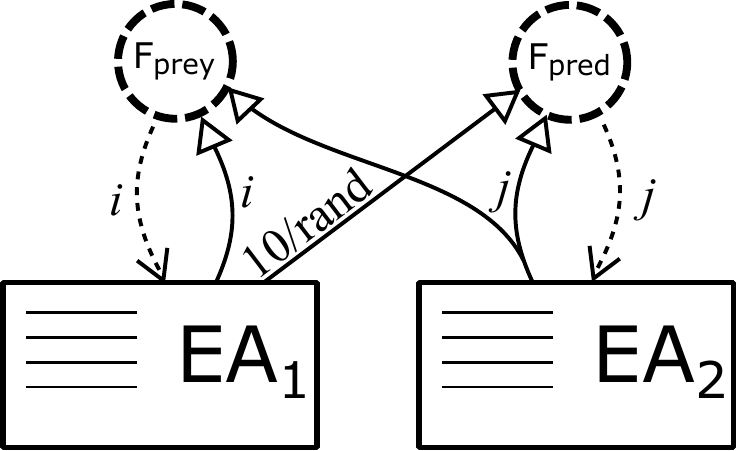}
					\label{sfig:comparative-rand}
				}
				\hspace{0.025\textwidth}
				\renewcommand{\thesubfigure}{f}
				\subfloat[
					The same system as depicted in \ref{sfig:comparative-coop} that is more clearly co-operative (though still subtly different from \ref{sfig:coop-multi}).
				]{
					\includegraphics[width=0.19\textwidth,keepaspectratio=true]{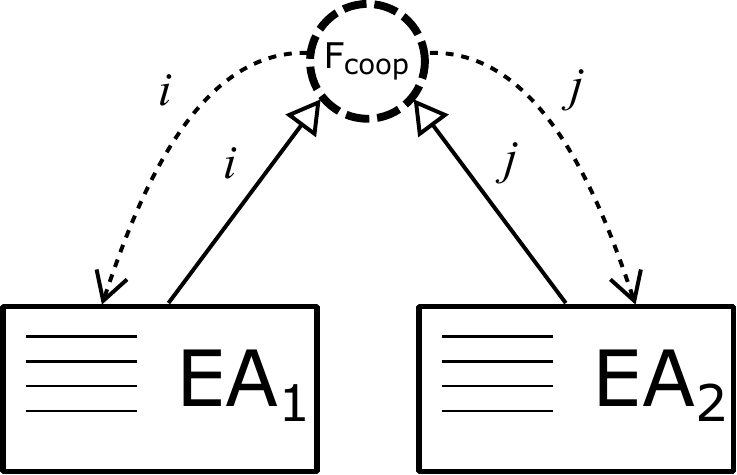}
					\label{sfig:coop-split}
				}
				\caption{Examples of edge labels.}
				\label{fig:multi}
			\end{figure}

					It should be noted that these categories of edge labels were developed while considering currently existing co-evolutionary and multi-population systems, and are likely to be far from exhaustive. It is almost unquestionable that extensions to the edge label notation will occur in future work as more use cases are considered.

					\subsubsection{Examples}
					\label{sssec:multi-examples}
					The utility of edge labels becomes apparent when considering the mpEAds in Figure \ref{fig:multi}, all of which are some variation on two-population co-evolution (as seen in Figures \ref{sfig:pred-prey} and \ref{sfig:coop}).

					Figures \ref{sfig:pred-prey-multi} to \ref{sfig:comparative-rand} describe variations on Figure \ref{sfig:pred-prey}. The first example, Figure \ref{sfig:pred-prey-multi}, is a common implementation of the predator/prey model, in which each predator is tested against all prey, and each prey is tested against all predators for their respective fitness values. This is computationally expensive (being $O(n^2)$), and subsequently,  variants using less than the full population are common.

					Figure \ref{sfig:pred-prey-multi-single} represents such a system, in which each individual from one population is paired with some other individual from the other population for evaluation. The method by which the other individual is selected is left unspecified. This individual could be randomly chosen, be the most fit, or selected by some other algorithm.

					In Figure \ref{sfig:comparative-rand}, which represents an asymmetrical approach to predator/prey (used only for demonstration purposes -- so far as we are aware, no such system exists in the literature). In this example, each predator is compared against ten prey, using the specified selection mechanism (drawing them at random). The prey, meanwhile, is evaluated in a fixed 1:1 pairing with a given predator.

					Figure \ref{sfig:coop-multi} depicts the simplest co-operative system that can be inferred from Figure \ref{sfig:coop}, using a similar 1:1 pairing of individuals from each of the populations to that seen in the predator/prey exaple in Figure \ref{sfig:comparative-rand}. This structure, though outwardly very different, is actually quite similar to that seen in Figure \ref{sfig:comparative-coop}, which, despite its superficial resemblance to the various predator/prey systems, is actually co-operative due to the 1:1 mapping between $i$ and $j$ for evaluation and the use of the same objective function for the evaluations.

					There is, however, a very subtle difference between Figures \ref{sfig:coop-multi} and \ref{sfig:comparative-coop}: in \ref{sfig:coop-multi}, the same information is sent to both populations (being stored at locations $i$ and $j$), whereas, in \ref{sfig:comparative-coop}, the information sent to both populations may not necessarily be the same. Finally, Figure \ref{sfig:coop-split} actually depicts the same functional system as \ref{sfig:comparative-coop}, as the different information streams are depicted separately rather than coming from the same source.

					\subsubsection{Examples of Greater Complexity}
					\label{sssec:multi-complex}
\begin{figure}
				\subfloat[
					A variation on predator/prey in which genotypes are decoded into phenotypes before being considered together.
				]{
					\includegraphics[width=0.19\textwidth,keepaspectratio=true]{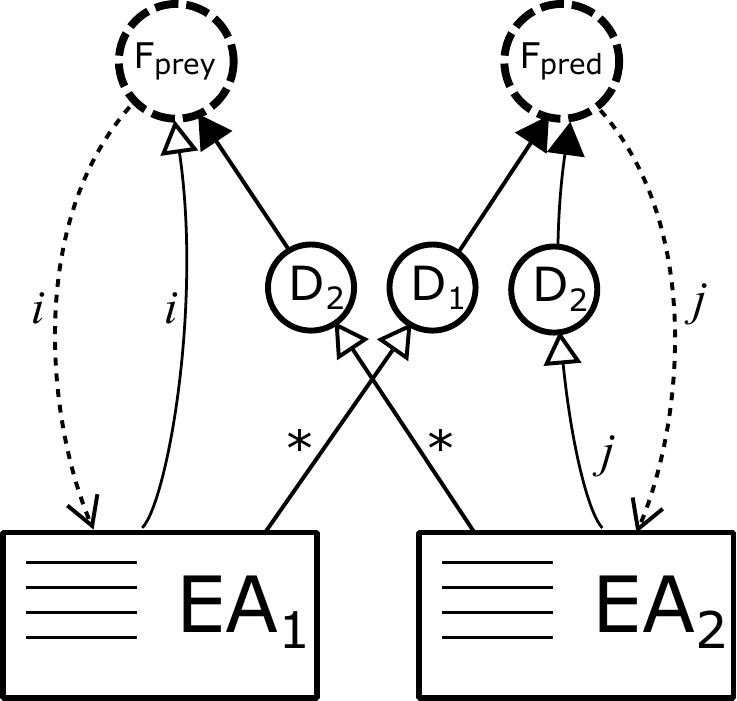}
					\label{sfig:pred-prey-decoded}
				}
				\hspace{0.025\textwidth}
				\subfloat[
					Predator/prey extended to include scavengers\cite{de2012co}.
				]{
					\includegraphics[width=0.19\textwidth,keepaspectratio=true]{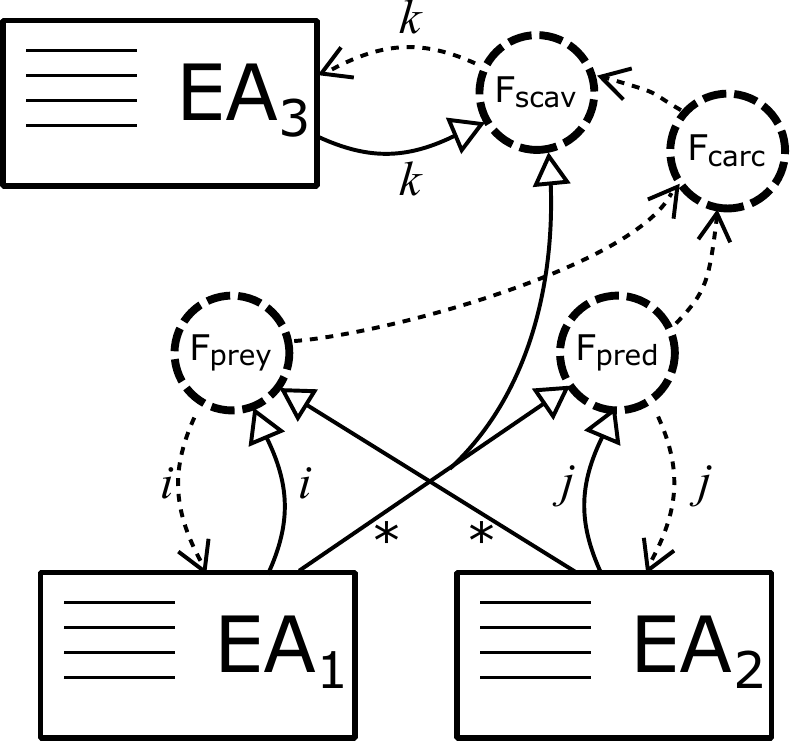}
					\label{sfig:pred-prey-scav}
				}
				\caption{Examples of more-advanced systems using edge labels.}
				\label{fig:multi-advanced}
			\end{figure}

					Figure \ref{sfig:pred-prey-decoded} depicts a variation on predator/prey in which the raw genotypes are first decoded into phenotypes before being used for standard predator/prey evaluation. It should be noticed that the borders of the decoding functions are solid, as the output of the computation nodes is phenotypic (and therefore genetic), which is represented using solid lines. The predator requires information from both the prey population and the predator population to be decoded before it can be evaluated, while the prey population only requires the predator to be decoded, while the genotypes of the prey themselves are acted upon directly.
					
					In Figure \ref{sfig:pred-prey-scav}, a third population is introduced to model a more general version of predator/prey based on the work of de Boer, Folkert, and Hogeweg\cite{de2012co}. The third population is composed of ``scavengers'' whose fitness is dependent on the fitnesses of both predator and prey, but which do not affect the fitness of either. The scavengers, much like their biological counterparts, exist only to ``pick up the scraps'' after the predator and prey have finished evaluating each other. In this system, the scavengers require the prey genotype to establish ``edibility''; after all, if the scavenger cannot digest the prey, it will go hungry!

					In this example, in addition to the three objective functions (one per population), there is a fourth function, F\textsubscript{carc}, where ``carc'' is an abbreviated form of ``carcass''. This function takes the evaluated fitnesses of each predator/prey pair and produces a ``fitness value'' for the consumption of the scavenger's objective function. This value, in effect, represents whether a given prey was actually killed (and can therefore be consumed), as well as how much of the prey is left over and can be used by the scavenger (hence ``carcass'', a term not used in the original paper\cite{de2012co}). The genetic information of the prey is also forwarded to the scavenger, in order to establish edibility as described above. It should be noticed that the mpEAd clearly represents all aspects of this process, as well as indicating the asymmetrical role of the scavenger population.

				\subsection{Inset Arrowheads}
\begin{figure}
			\centering
			\includegraphics[width=0.4\textwidth,keepaspectratio=true]{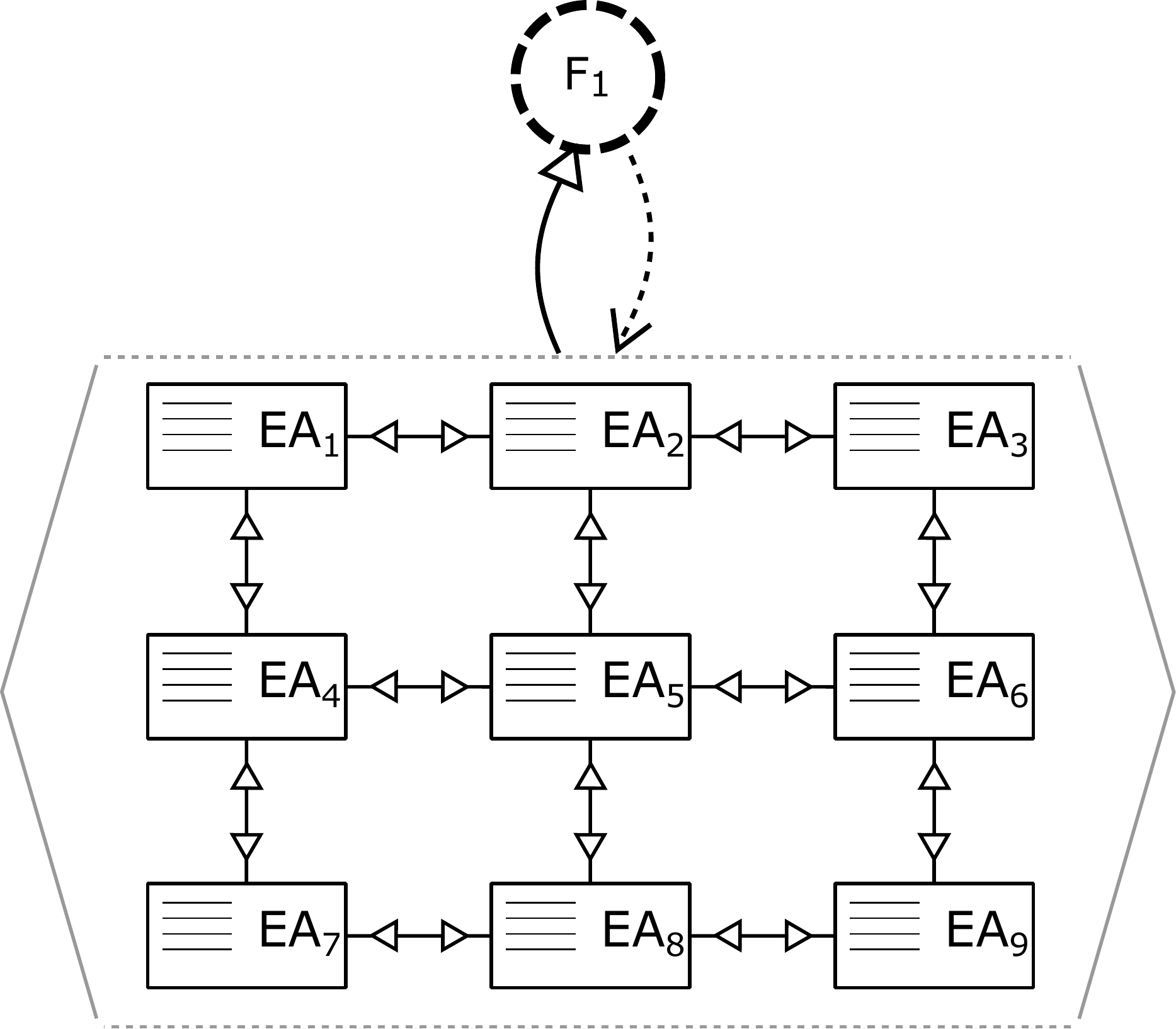}
			\caption{A small island-model system using a 3x3 grid of populations.}
			\label{fig:small-grid}
		\end{figure}

					Migration between populations is a common feature of multi-population systems and it would be remiss to exclude it from mpEAd. Yet, migration presents an apparent quandary: at first glance, it appears to break with the ``information flow'' model, as an actual individual is being transported rather than formless information. On deeper reflection, it becomes obvious that migration can be modelled as a transfer of genetic information followed by a state change in the receiving population. There is, however, a distinction to be made between the arrival of genetic information to be added to the population versus the arrival of genetic information to be incorporated into existing individuals in that population. The first, obviously, is migration. The second is lesser-known, although still existing in the literature, where it is known as hierarchical composition\cite{watson2002compositional}. In every other instance in mpEAd, arrowheads that touch a population node affect individuals within that population. Consequently, a genotypic arrowhead that touches the node border maps more closely to hierarchical composition (which changes the individual) than migration (which adds an individual to the population). Migration, then, is modelled using inset arrows located along the edge in question (as seen in Figure \ref{fig:small-grid}).

		\begin{figure}[b]
			\centering
			\includegraphics[width=0.48\textwidth,keepaspectratio=true]{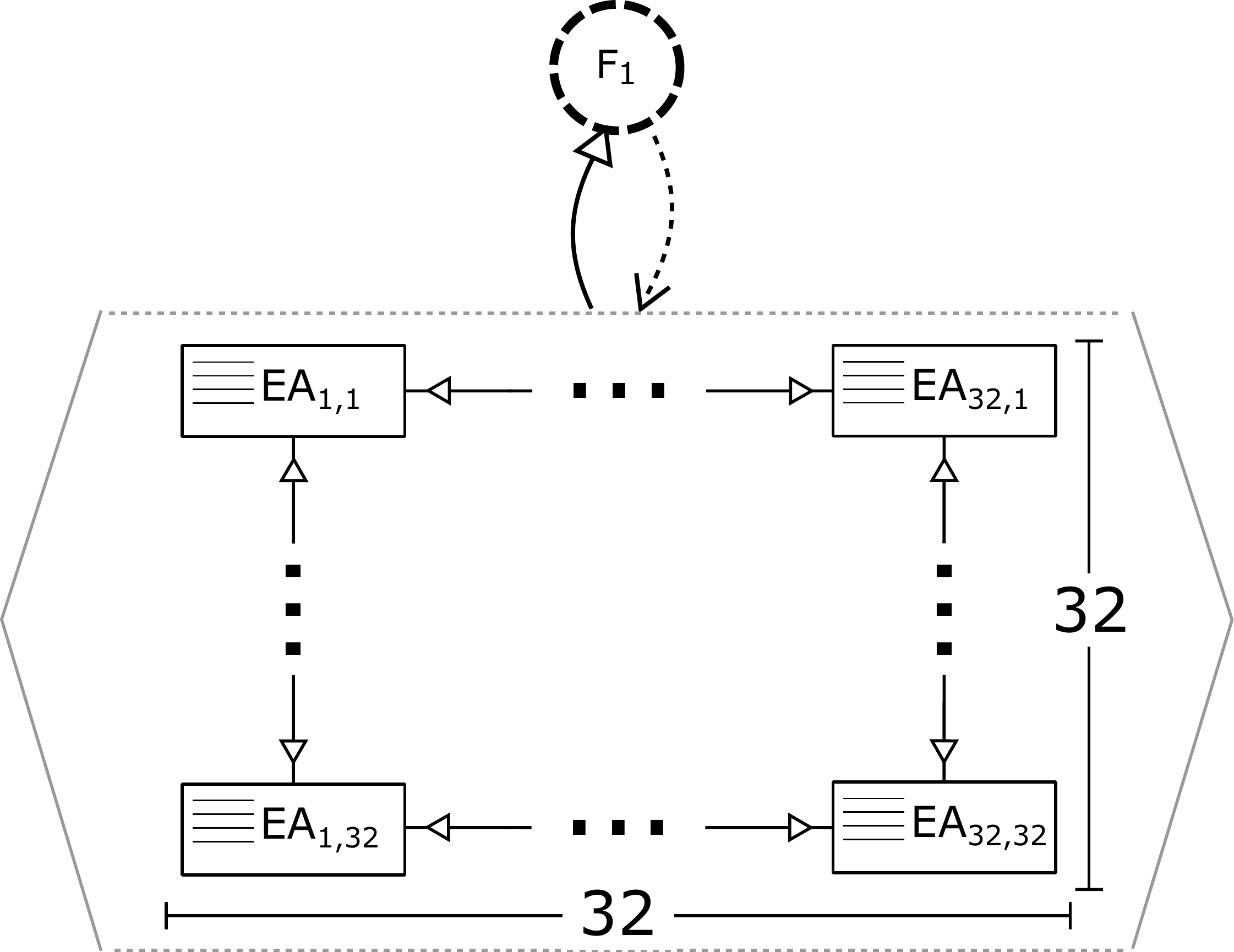}
			\caption{A much larger island-model system, using a 32x32 grid of populations.}
			\label{fig:big-grid}
		\end{figure}

				\subsection{Macro Boxes and Ellipsis Notation}
					A macro box is, much like in a programming language, a simple shorthand for repeating elements. This is typically used in mpEAd to depict the same evaluation being used independently for multiple populations (this being distinct from a single evaluation shared by multiple populations, which implies co-evolution).

					An element that originates arrows ending at the border of the macro box is duplicated and connected to every element within the box, while elements whose arrows do not end at the macro box border are not duplicated. Macro boxes are are depicted using a elongated gray hexagon in which the long face is dashed and the remaining faces are solid. A simple example of macro box usage can be seen in Figure \ref{fig:small-grid}, in which a traditional island-model EA is shown with migration occurring across a 3x3 grid of populations. While macro boxes have many additional uses and properties, a full discussion of these is outside the scope of this paper.

					The ellipsis (\ldots) is used analogously to abbreviating text. It indicates the presence of many more identical components of a system, the actual drawing of which would be difficult or unwieldy. The ellipsis is always accompanied by a horizontal bar with vertical ends and an integer (value greater than 2) that indicates the total number of elements, including those actually drawn. An example of this usage is given in Figure \ref{fig:big-grid}, in which a 32x32 migratory grid is depicted in a compact form.

\begin{figure}
			\includegraphics[width=0.5\textwidth,keepaspectratio=true]{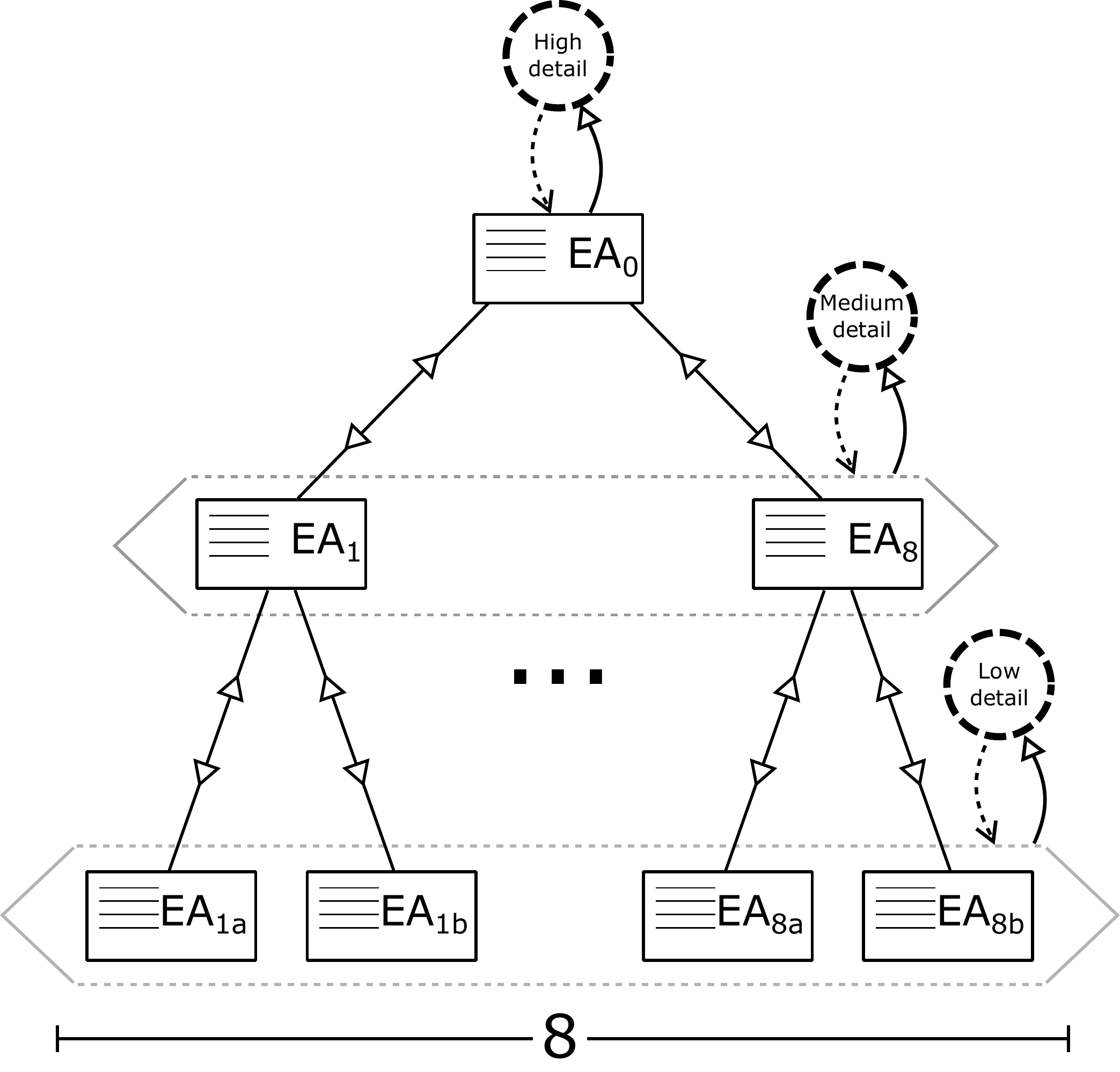}
			\caption{An extended version of the hierarchical GA created by Sefrioui and P\'eriaux\cite{sefrioui2000hierarchical}.}
			\label{fig:sefrioui}
		\end{figure}

					The full power of the combination of these two concepts can be found in Figure \ref{fig:sefrioui}, which depicts an extended version of the hierarchical GA created by Sefrioui and P\'eriaux\cite{sefrioui2000hierarchical}. This system is used to find solutions for an problem for which the objective function is computationally expensive to evaluate, but can be approximated using functions that require less computational effort at the expense of precision. Sefrioui and P\'eriaux first use a number of coarsely-evaluated populations to search for promising solutions, which are then passed upwards in order to be more finely evaluated. They also allow solutions to be passed down, in order to assist the lower-level populations and keep them ``current''. While the original version takes the form of a binary tree of populations, this variant adds additional subtrees below the ``high detail'' node, resulting in a system of 25 nodes rather than the original 7. Such a system would be able to more effectively cover the entire search space than the original model, potentially accelerating the process of locating a solution to a complex problem.

		\section{Conclusion}
		\label{sec:conc}
			The mpEAd formalism is a graphical notation designed to permit the depiction of large, complex multi-population EA systems. Designed with the goals of being as intuitive, consistent, distinctive, and simple as possible, mpEAd is a powerful modelling tool for systems often considered too complex to describe clearly. Even with a system as small as two populations, we have seen that, through the use of mpEAd, it is possible to envision not only existing systems, but to diagram and reason about systems not in the literature in an easy and clear way. In the future, even more elements may be added to mpEAd as EC itself grows and matures, necessitating unenvisioned interactions and relationships. Ultimately, it could be possible to develop an ``mpEAd IDE'', where complex systems are created visually before being rendered down to source code automatically. While these pursuits remain in the future, it is clear that mpEAd has practical applications in the here and now, depicting the exceedingly complex in a simple manner.

	\printbibliography
\end{document}